\documentclass{article}

\usepackage{microtype}
\usepackage{graphicx}
\usepackage{subcaption}
\usepackage{booktabs} 
\usepackage{multirow}
\usepackage{xcolor}
\usepackage{enumitem}
\usepackage{tcolorbox}

\usepackage{hyperref}

\usepackage[preprint]{icml2026}

\usepackage{amsmath}
\usepackage{amssymb}
\usepackage{mathtools}
\usepackage{amsthm}

\usepackage[capitalize,noabbrev]{cleveref}

\theoremstyle{plain}

\theoremstyle{definition}

\theoremstyle{remark}

\usepackage[textsize=tiny]{todonotes}

\usepackage{xspace}

\icmltitlerunning{Balancing Faithfulness and Performance in Reasoning via Multi-Listener Soft Execution}
\newcommand{\ie}{i.e., }

\newcommand{\editedcontent}[1]{#1}

\newcommand{\myparagraph}[1]{\textbf{#1}\hspace{0.4em}}

\newcommand{\method}[0]{\textsc{REMuL}\xspace}

\begin{document}

\twocolumn[
  \icmltitle{Balancing Faithfulness and Performance in Reasoning \\
  via Multi-Listener Soft Execution}



  \icmlsetsymbol{equal}{*}

  \begin{icmlauthorlist}
    \icmlauthor{Nithin Sivakumaran}{unc}
    \icmlauthor{Shoubin Yu}{unc}
    \icmlauthor{Hyunji Lee}{unc}
    \icmlauthor{Yue Zhang}{unc}
    \\
    \icmlauthor{Ali Payani}{cisco}
    \icmlauthor{Mohit Bansal}{unc}
    \icmlauthor{Elias Stengel-Eskin}{ut}
  \end{icmlauthorlist}

  \icmlaffiliation{unc}{UNC Chapel Hill}
  \icmlaffiliation{cisco}{Cisco}
  \icmlaffiliation{ut}{University of Texas at Austin}

  \icmlcorrespondingauthor{Nithin Sivakumaran}{nsivaku@cs.unc.edu}

  \icmlkeywords{Machine Learning, ICML}
  \vskip 0.3in
]



\printAffiliationsAndNotice{}  

\begin{abstract}
  Chain-of-thought (CoT) reasoning sometimes fails to faithfully reflect the true computation of a large language model (LLM), hampering its utility in explaining how LLMs arrive at their answers.
  Moreover, optimizing for faithfulness and interpretability in reasoning often degrades task performance.
  To address this tradeoff and improve CoT faithfulness, we propose Reasoning Execution by Multiple Listeners (\method{}), a multi-party reinforcement learning approach. 
  \method{} builds on the hypothesis that reasoning traces which other parties can follow will be more faithful. 
  A speaker model generates a reasoning trace, which is truncated and passed to a pool of listener models who ``execute'' the trace, continuing the trace to an answer. 
  Speakers are rewarded for producing reasoning that is clear to listeners, with additional correctness regularization via masked supervised finetuning to counter the tradeoff between faithfulness and performance.
  On multiple reasoning benchmarks (BIG-Bench Extra Hard, MuSR, ZebraLogicBench, and FOLIO), \method{} consistently and substantially improves three measures of faithfulness -- hint attribution, early answering area over the curve (AOC), and mistake injection AOC -- while also improving accuracy.
  Our analysis finds that these gains are robust across training domains, translate to legibility gains, and are associated with shorter and more direct CoTs.\footnote{Code: \href{https://github.com/nsivaku/remul}{\texttt{https://github.com/nsivaku/remul}}}
\end{abstract}

\begin{figure*}[ht]
  \begin{center}
    \centerline{\includegraphics[width=\textwidth]{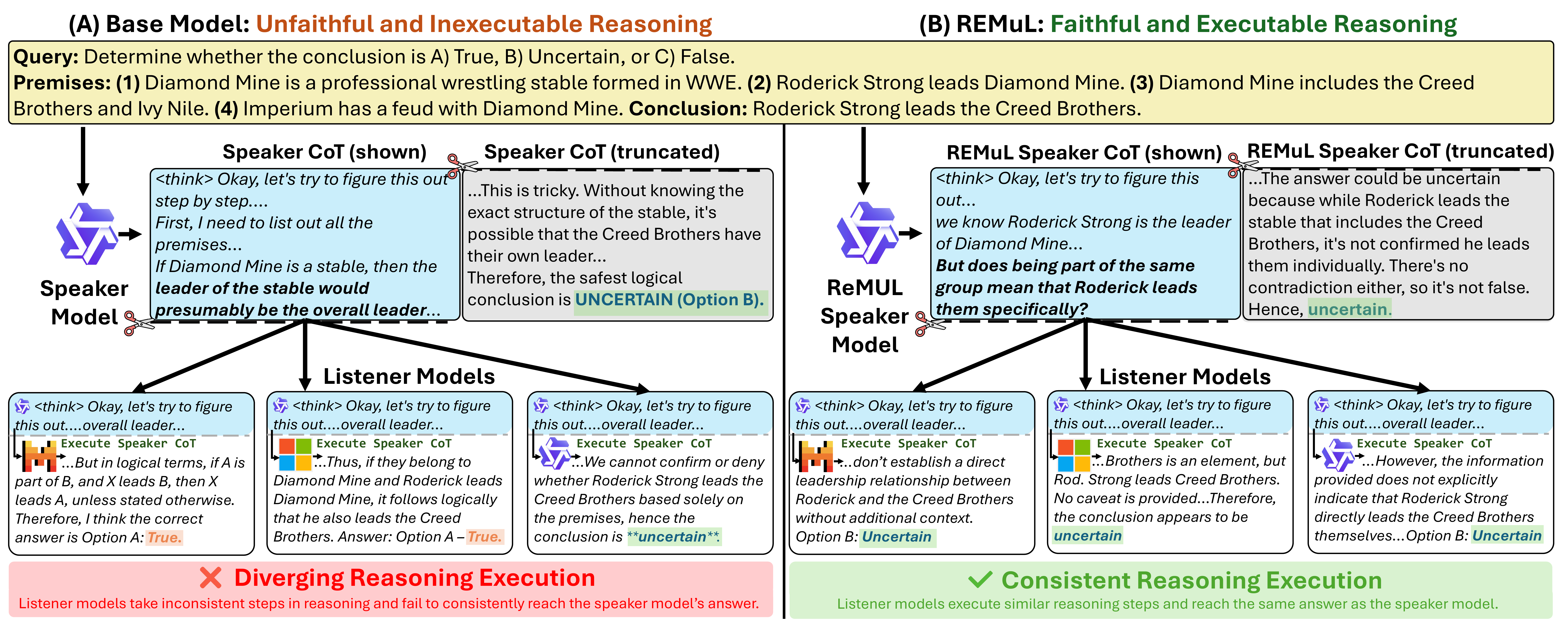}}
    \caption{
      Existing reasoning models are prone to producing reasoning that is unfaithful and hard to follow. 
      Thus, when given a reasoning chain prefix (blue) from a base speaker model (A), 
      listener models are often unable to reproduce the speaker's final decision, i.e., their soft execution suffix (white) diverges. 
      \method{} rewards speakers for inducing consistent reasoning in the listener pool.
      With \method{}, the speaker's reasoning is more faithful and easier for the listeners to follow consistently, with all listeners reaching the same conclusion as the speaker.
      Note that reasoning here is elided, and the shown prefix and truncated suffix are not by necessity the same length.
    } 
    \label{fig:fig1}
  \end{center}
\end{figure*}

\section{Introduction}

While current reasoning large language models (LLMs) produce chain-of-thought (CoT) outputs that ostensibly reflect their reasoning, it remains unclear to what extent these outputs faithfully represent the model's true computational process \citep{lanham2023measuring, barez2025chain, chen2025reasoning}. 
This gap limits the reliability of LLMs and makes verifying their answers challenging, especially in domains (e.g., medicine, law, scientific research) where users require not just an answer but a faithful explanation of how that answer was reached \citep{sitaraman2025ai, mashraqi2022current, rajpurkar2022ai, resnik2025ethics, schwarcz2025ai}. 
Moreover, a line of prior work points to a tension between task performance and reasoning faithfulness, with improvements to accuracy often coming at a cost to reasoning faithfulness or interpretability \citep{arrieta2020explainable, kenny2024regulation, mcmillan2025towards}.  

In the context of LLM reasoning, the risk of unfaithful reasoning is compounded by reinforcement learning (RL)-based approaches, which typically reward the accuracy of the final answer but not the faithfulness of the model's reasoning \cite{lambert2024tulu}.
Indeed, past work has found that outcome-only RL can lead to unfaithful or uninterpretable outputs \cite{guo2025deepseek, kirchner2024prover, jose2025reasoning}. 
Unlike correctness, which can be objectively defined, we argue that faithfulness is a multi-party property: 
given a speaker and a listener, a faithful explanation is one which enables the listener model to come to the same conclusion as the speaker, without access to the speaker's answer.
For example, in \cref{fig:fig1}(A), we see that when we truncate a reasoning chain from a speaker model to exclude its final few steps and its answer, listener models use the same reasoning prefix to come to different conclusions, indicating that the speaker's reasoning prefix does not necessarily clearly lead to their final answer.
We frame this as a question of \emph{reasoning executability}: a reasoning chain is faithful if it can be executed by a similarly capable listener 
to recover the same conclusion, without access to the speaker’s answer; execution here refers to a ``soft execution'', where a prefix 
of the speaker's reasoning chain is provided to the listener, who follows it to its conclusion. 
Using this framing, we propose \underline{R}easoning \underline{E}xecution by \underline{Mu}ltiple \underline{L}isteners (\method{}), which trains models for both faithfulness and correctness simultaneously.

Concretely, we train a speaker policy using Group Relative Policy Optimization (GRPO) \citep{guo2025deepseek} where the speaker generates a reasoning trace, which we then truncate to produce a prefix.
This is illustrated in \cref{fig:fig1}, where the speaker CoT is cut into two portions, prefix and suffix.
The prefix is then passed to multiple independent listener models, while the truncated suffix is not shown.
Each listener begins execution from the prefix and produces its own answer. 
As shown in \cref{fig:method} (top), the speaker is rewarded when the listener agents converge to the same answer, with higher rewards for stronger consensus. 
Note that this process does not necessarily require the listener models to converge to a correct answer, but only to match the speaker's answer. 
Indeed, we find that simultaneously aiming to reward for consistent listener reasoning execution and answer correctness often results in degraded performance for both metrics.
Specifically, consistent with prior work \citep{arrieta2020explainable, kenny2024regulation, mcmillan2025towards}, we find that increasing faithfulness alone comes at a cost to correctness.
To offset this, we incorporate correctness training into \method{}.
Finding that jointly rewarding correctness and faithfulness via GRPO tends to lead to competing rewards and suboptimal performance, we introduce a mixed method which optimizes for faithfulness via RL and optimizes for correctness using supervised finetuning (SFT) with a LoRA \citep{hulora} adapter, shown in \cref{fig:method} (bottom). 

After training on data from BIG-Bench Hard \citep[BBH;][]{suzgun2023challenging}, we evaluate \method{} on challenging reasoning problems from BIG-Bench Extra Hard \citep{kazemi2025big}, logic puzzles from ZebraLogicBench \citep{linzebralogic}, multi-step logical deduction puzzles from MuSR \citep{spraguemusr}, and first-order logic questions from FOLIO \citep{han2024folio}. 
We measure faithfulness using intervention-based area over the curve (AOC) metrics \citep{lanham2023measuring} as well as hint usage \citep{chen2025reasoning}, which tests whether model outputs faithfully verbalize changes in answers due to user-provided hints.
Here, we find that our method improves hint attribution by up to 3.1\% on Qwen3 and 2.7\% on Phi4, while also boosting accuracy by up to 2.1\% and 3.2\%, respectively. 
We additionally observe gains of up to 4.6 AOC points for early answering from truncated CoT (a faithfulness measure introduced by \citet{lanham2023measuring}).

In our analysis, we explore \method{}'s impact on several dimensions of model behavior and justify key design choices. First, we find that our method improves \textit{legibility} \citep{emmons2025pragmatic} in model outputs, meaning that the produced reasoning chains are rated more clear and readable. We also perform an ablation study on our speaker-listener setup to demonstrate the importance of having multiple diverse listeners during training. Without either, we see diminished gains in both accuracy and faithfulness measures. 
We also observe a conflict in correctness and faithfulness reward signals when jointly optimized under a single reward,
explaining the tradeoff observed under joint optimization and further motivating our multi-loss approach, which optimizes for correctness and faithfulness via distinct mechanisms.
Finally, we study the impact of using in-domain data and find that while it does provide substantial boosts in accuracy, it does not necessarily result in gains to faithfulness. Our key contributions are summarized as follows:
\begin{itemize}[noitemsep,topsep=0pt,leftmargin=*]
    \item We propose \method{}, a novel multi-party RL framework that trains models to produce faithful reasoning by rewarding speaker models for generating reasoning that is consistently executable by a diverse group of listener models, while simultaneously regularizing for correctness. 
    \item We show consistent gains on different faithfulness measures -- hint attribution, AOC for early answering after CoT truncation, and AOC for mistake injection -- while maintaining or improving task accuracy on multiple challenging reasoning benchmarks.
    \item Our analysis indicates that \method{}'s gains translate to legibility, and we show that \method{}'s improved performance correlates with more concise and linear reasoning.
\end{itemize}

\section{Method}
\label{method}

\subsection{Problem Setup}
Given an input prompt $x$, a speaker model $S$ generates an explicit reasoning trace $t$ and a final answer $a$, yielding an output $y$.
Our goal is to train $S$ to produce more faithful reasoning. 
More specifically, the produced trace $t$ should better reflect the process required to reach the answer $a$. 
We formalize this computation via the notion of soft executability by a listener model $L$.

\subsection{Cross-model Soft Reasoning Execution}
\label{sec:cross_model}
More formally, let $t = \langle t_1, t_2, \ldots, t_n\rangle$ be a reasoning chain with $n$ steps, leading to answer $a$. 
We define a soft execution of $t$ as $M(x, t_{1,m})=a'$, where $M$ is a model, $x$ is an input prompt, $t_{1,m}$, $m \leq n$ is a partial reasoning chain, and $a'$ is the listener model response.\footnote{We refer to this execution as ``soft'' to distinguish it from programmatic execution, e.g., \citet{lyu2023faithful}.}
Note that in this formulation, $M$ completes the remaining reasoning steps to reach a potentially different answer $a'$.
We define our faithfulness reward as $\mathbb{I}[L(x, t_{1,m}) = S(x, t)]$ (or, equivalently, $\mathbb{I}[a' = a$]) i.e. the degree to which a listener model $L$ reaches the same answer as the original speaker reasoning model $S$. 
This process is illustrated in \cref{fig:fig1}, where the speaker model's partial chain is passed to multiple listeners.

\begin{figure}[t]
    \centering
    \includegraphics[width=\linewidth]{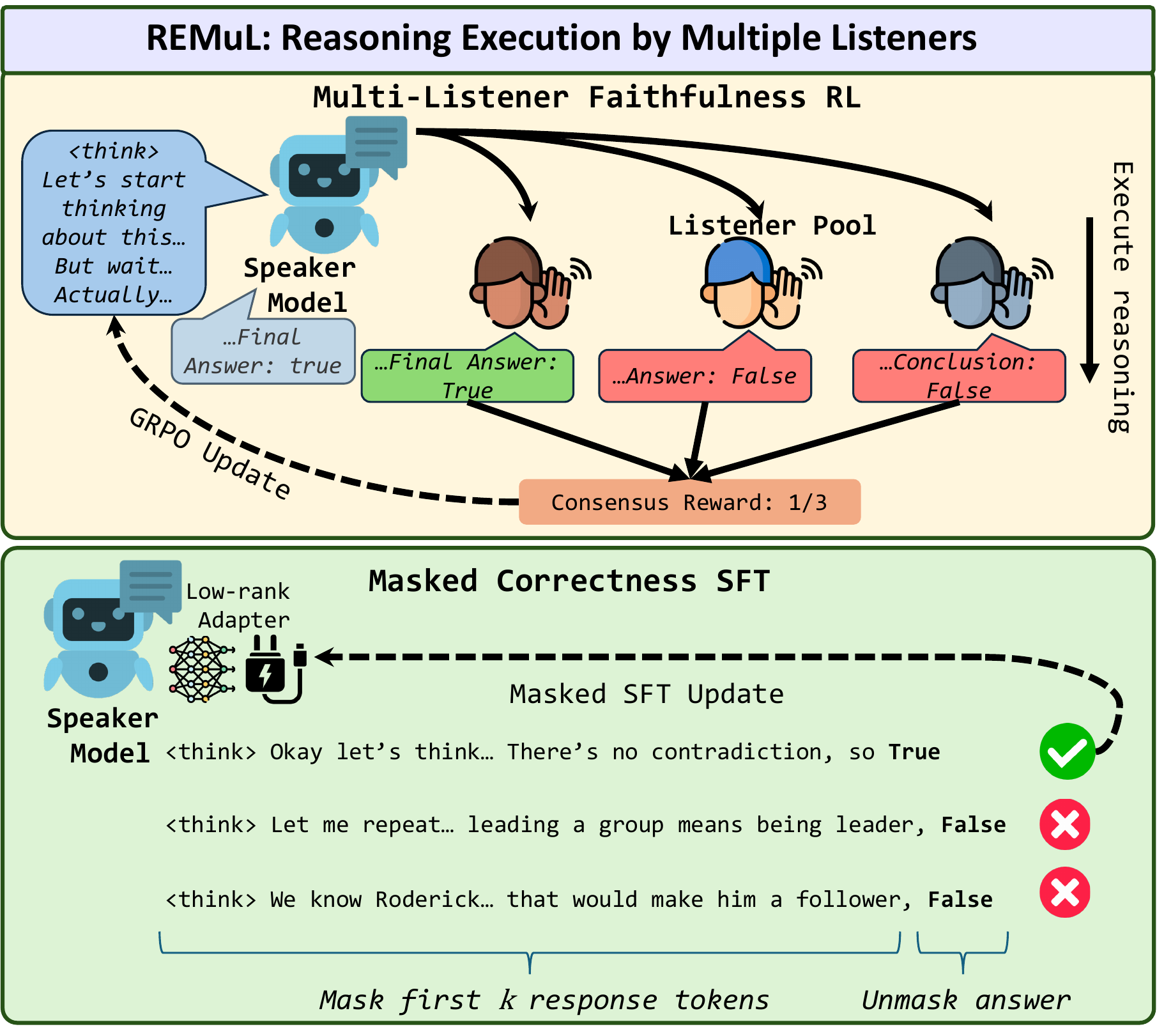}
    \caption{\method{} consists of two components: (Top) A speaker-listener reasoning execution reward, where listeners execute reasoning prefixes from a speaker, who is rewarded for listener consensus. The speaker's final answer is only used for reward computation and not seen by the listeners. (Bottom) A masked supervised finetuning step to maintain correctness via a LoRA adapter, with loss computed only on answer tokens.}
    \label{fig:method}
\end{figure}

As illustrated in \cref{fig:method} (top), \method{} uses cross-model execution as a training signal.
We generate an initial response to input $x$ using $S$ and generate reasoning trace $t$ and final answer $a$. 
We then truncate $t$ to 25\%, 50\%, and 75\% of its length (split on newlines) to get $T' = \{t_1, t_2, t_3\}$. 
Then, given a set of listener models $\mathcal{M}$, for each listener model $L_i \in \mathcal{M}$, we obtain a finished reasoning trace starting from $t_j \in T'$ and final listener answer of $a_j^i$. 
We compute a matching reward across the pool by the formula:
\begin{equation}
r_{match} = \sum_{L_i\in\mathcal{M}} \sum_{t_j\in T'} \mathbb{I}\!\left[a_j^i = a\right]
\label{eq:match_reward}
\end{equation} 
Based on this continued generation process, we reward the speaker policy using $r_{match}$. The main assumption is that if the speaker policy's reasoning is logically sound, listener models should be more likely to converge to the same answer when continuing from the trace.
Note that when training with $r_{match}$ we use full-rank updates.

\subsection{Incorporating Correctness}
Due to observed accuracy tradeoffs (see \cref{sec:tradeoff}) when optimizing only for faithfulness, we test a variety of methods to improve correctness results during training.

\paragraph{Balanced Rewards.}

We train $S$ with a reward that combines matching under cross-model execution and correctness term based on ground truth label $g$. Under the balanced reward, the full reward function is as follows:
\begin{equation}
    r_{bal}(x,y) = r_{match}+\lambda\,\mathbb{I}[a = g],
    \label{eq:baL_reward}
\end{equation}
where $\lambda$ is defined as the number of the listener models (\ie $\;|\mathcal{M}|$). 
This balanced reward is then used when doing GRPO on our training data.

\paragraph{Correctness Supervised Finetuning.}
\label{sec:correctness_sft}

We explore a supervised finetuning (SFT) strategy to mitigate accuracy drops while optimizing for faithfulness.
In this approach, as illustrated in \cref{fig:method} (bottom), we add a standard SFT loss to supervise only the final answer tokens while masking the reasoning trace, similar to \citet{fang2025thinkless}. 
This isolates the correctness signal to the answer channel and avoids directly altering the reasoning trace. 
In other words, our loss is given by:
\begin{equation}
\mathcal{L}_{\text{ans}}(\theta) \;=\; - \sum_{i \in \mathcal{A}} \log p_{\theta}\!\left(y_i \mid x, y_{<i}\right),
\end{equation}
where $\mathcal{A}$ indexes the token positions corresponding to the final answer and token positions belonging to $t$ receive zero loss.
Intuitively, this preserves the reasoning behaviors learned during faithfulness training (since we do not explicitly force a particular chain-of-thought), while still correcting the final decision boundary. 

\subsection{Training Procedure}
We first optimize the original speaker model $S$ using Group Relative Policy Optimization (GRPO) \cite{guo2025deepseek}.
For each prompt $x$, we sample a group of $G$ outputs $\{y_i\}_{i=1}^G$ from $S$; then, for each $y_i$ we apply truncation, pass the truncated traces through the population of listeners $\mathcal{M}$, and compute the reward $r_{match}$. When using a balanced reward that accounts for correctness, we instead use $r_{bal}$, an augmented version of $r_{match}$ defined in \cref{eq:baL_reward}.
This reward is used to calculate relative advantages and increase the likelihood of higher-advantage outputs, iteratively teaching the speaker policy $S$ to output more executable and faithful traces.

Separately, we finetune LoRA parameters using the masked correctness loss on top of the faithfulness-trained speaker checkpoint. 
We then merge the resulting trained LoRA adapter back into the original speaker checkpoint. 
Empirically, we find that this approach is especially effective for recovering accuracy while maintaining the behaviors induced by our multi-party training.
Note that at test-time, only a single model (the speaker) is required, making the requirement for multiple models a fixed, training-time cost. 

\section{Experiments and Results}
\label{results}

\subsection{Experimental Setup}
\label{sec:exp_setup}
\paragraph{Training data.}
We train on a subset of Big-Bench Hard (BBH) consisting of five tasks chosen because they reliably benefit from chain-of-thought (CoT) reasoning \citep{suzgun2023challenging}: 
(1) \emph{Logical deduction -- 5 objects}, (2) \emph{Navigate}, (3) \emph{Temporal}, (4) \emph{Sports understanding}, and (5) \emph{Shuffled objects}.
We train Qwen3-14B \citep{yang2025qwen3} and Phi4 \citep{abdin2025phi} with thinking enabled for 3 epochs on this subset of 1250 QA pairs.

\paragraph{Evaluation datasets.}
We evaluate on four multi-step reasoning benchmarks:
Big-Bench Extra Hard \citep[BBEH;][]{kazemi2025big}, ZebraLogicBench \citep[ZLB;][]{linzebralogic}, FOLIO \citep{han2024folio}, and MuSR \citep{spraguemusr}.
For BBEH, we filter the benchmark to include only multiple-choice questions to match the training data and other evaluation setups, yielding 120 questions. For MuSR, we evaluate on the 250 question murder mysteries split of the dataset. For ZLB and FOLIO, we use the full multiple-choice splits with 3,259 and 202 question respectively. All four benchmarks require multi-step inference to reach the final answer and are designed to test models' reasoning capabilities.

\paragraph{Implementation details.}
Our training procedure follows the multi-party GRPO scheme described in Section~\ref{method}, where listener models provide a matching-based learning signal via cross-model execution. To construct truncated reasoning traces, we split the model's reasoning (contained between the \texttt{<think>} tag) by newline characters 
and take the first $p$\% of lines as the truncated reasoning trace $t_j$.
Listener models $\mathcal{M}$ are prompted to continue from this truncated reasoning and produce a final answer, which we use to compute matching rewards. The listener models we use are Ministral-3-14B-Reasoning \citep{liu2026ministral}, Phi-4-reasoning-plus 14B \citep{abdin2025phi}, and Qwen3-14B \citep{yang2025qwen3}.
Across all methods, we keep prompts and answer formatting consistent within each dataset and evaluate each model under the same truncation and answer parsing rules. Details about individual model configurations are provided in \cref{app:model_configs}.
Prompts are given in \cref{app:prompts}.

\subsection{Metrics}
\label{sec:metrics}

We report end-task performance in terms of accuracy, as well as the following faithfulness metrics: 

\myparagraph{Hint Usage.}
We measure faithfulness using the hint-injection protocol from \citet{chen2025reasoning}.
For each example, we evaluate the model on the original prompt and a modified prompt containing an explicit hint. We use the sycophantic hint outlined in \citet{chen2025reasoning} (e.g., \emph{``Hint: I think the answer is Option A: True. What do you think?"}).
If the model's final answer changes between the two prompts, we check whether the model attributes the change to the hint in its explanation (rather than presenting the new answer as independently derived). The intuition is that if a model changed its answer from its original response without referencing the hint, then it is not faithfully expressing its decision making process, which depends on the hint. Inherently, this metric can only consider examples where the model changed its answer.
We report hint usage as the percentage of changed-answer cases where the model explicitly cites the hint as the reason for the change.  We detect hint usage by checking for a set of tokens listed in \cref{app:hint_tokens}.

\begin{table*}[t]
\centering
\caption{Accuracy and hint usage across different reasoning benchmarks. \method{} consistently improves hint usage over the correctness baseline while maintaining or improving accuracy, demonstrating an effective balance between task performance and faithful reasoning. 
We \textbf{bold} and \textit{italicize} the first and second-best results, respectively. 
Hint Optimized (grayed) is excluded from ranking as it is directly optimized for the hint usage test metric.
} 
\small
\begin{tabular}{llcccccccc}
\toprule
\multirow{2}{*}{\textbf{Model}} &
\multirow{2}{*}{\textbf{Method}} &
\multicolumn{2}{c}{\textbf{BBEH}} &
\multicolumn{2}{c}{\textbf{ZLB}} &
\multicolumn{2}{c}{\textbf{MuSR}} &
\multicolumn{2}{c}{\textbf{FOLIO}} \\
\cmidrule(lr){3-4} \cmidrule(lr){5-6} \cmidrule(lr){7-8} \cmidrule(lr){9-10}
& & \textbf{Acc} & \textbf{Hint} & \textbf{Acc} & \textbf{Hint} & \textbf{Acc} & \textbf{Hint} & \textbf{Acc} & \textbf{Hint} \\
\midrule
\multirow{7}{*}{\textit{Qwen3-14b}} 
& Original                               & 31.3 & 56.7 & 32.6 & 67.3 & \textit{65.8} & 76.4 & 75.3 & 63.0 \\
& MAT-Steer                          & 32.6 & 58.1 & 31.8 & 68.9 & \textbf{66.2} & 78.3 & 77.1 & 63.4 \\
& Faithfulness Only                  & 28.4 & \textbf{61.8} & 28.9 & \textbf{70.6} & 60.5 & \textbf{80.1} & 74.5 & \textbf{67.1} \\
& Correctness Only                   & \textbf{34.5} & 53.8 & \textbf{38.3} & 65.7 & 65.4 & 72.8 & \textbf{77.8} & 62.1 \\
& Balanced Rewards                   & 28.9 & 52.4 & 33.5 & 66.3 & 62.2 & 74.5 & 74.5 & 61.5 \\
& \textcolor{gray}{Hint Optimized}   & \textcolor{gray}{26.1} & \textcolor{gray}{71.2} & \textcolor{gray}{27.8} & \textcolor{gray}{79.5} & \textcolor{gray}{58.7} & \textcolor{gray}{87.6} & \textcolor{gray}{72.8} & \textcolor{gray}{75.4} \\
\cmidrule(lr){2-10}
& \method{}                          & \textit{33.3} & \textit{59.3} & \textit{35.7} & \textit{70.1} & \textit{65.8} & \textit{79.5} & \textit{77.3} & \textit{64.6} \\
\midrule
\multirow{7}{*}{\textit{Phi4-reasoning-plus}} 
& Original                               & 26.6 & 45.2 & 27.4 & 61.6 & 72.3 & 65.1 & 79.8 & 54.8 \\
& MAT-Steer                          & 26.9 & 46.0 & 26.9 & 62.2 & \textit{73.1} & 66.4 & 80.6 & 53.9 \\
& Faithfulness Only                  & 23.5 & \textbf{49.8} & 26.3 & \textbf{65.0} & 67.0 & \textbf{68.2} & 80.4 & \textbf{58.4} \\
& Correctness Only                   & \textbf{29.1} & 44.1 & \textbf{32.3} & 61.7 & \textbf{73.9} & 62.0 & \textbf{83.0} & 53.5 \\
& Balanced Rewards                   & 25.8 & 44.7 & 26.4 & 63.1 & 69.1 & 63.5 & 81.2 & 54.0 \\
& \textcolor{gray}{Hint Optimized}   & \textcolor{gray}{21.8} & \textcolor{gray}{53.4} & \textcolor{gray}{24.1} & \textcolor{gray}{70.8} & \textcolor{gray}{65.3} & \textcolor{gray}{73.0} & \textcolor{gray}{78.2} & \textcolor{gray}{65.3} \\
\cmidrule(lr){2-10}
& \method{}                          & \textit{28.2} & \textit{48.6} & \textit{30.6} & \textit{64.3} & 71.6 & \textit{67.7} & \textit{81.7} & \textit{56.2} \\
\bottomrule
\end{tabular}
\label{tab:main}
\end{table*}

\paragraph{AOC Metrics.}
\editedcontent{Here, we use the \textit{early answering} and \textit{adding mistakes} area-over-the-curve (AOC) metrics from \citet{lanham2023measuring}. 
This metric truncates CoT reasoning at various split-points. For the ``Truncated CoT Answering'' setting, we force the model to answer at each point, measuring the accuracy. The higher the AOC is, the more of the reasoning chain is required to correctly answer, i.e., an AOC of 1 indicates that the model cannot answer correctly until the full reasoning is achieved.
On the other hand, a lower AOC indicates that most of the reasoning is post-hoc rationalization, and that the answer is already inevitable early on in the reasoning chain. For the ``Adding Mistakes'' setting, we corrupt the reasoning chain by inserting errors at each split-point and allow the model to continue generating from the corrupted point onward. A higher AOC indicates that the model's final answer is more sensitive to mistakes in its reasoning, suggesting it is faithfully conditioning on its chain of thought. Conversely, a lower AOC suggests the model arrives at the correct answer despite the introduced errors, indicating that the reasoning is not being faithfully followed.
In our experiments, we measure accuracy at evenly spaced 20\% increments along the original reasoning chain.}

\subsection{Baselines}
We compare our trained models to the following baselines:
\begin{itemize}[noitemsep,topsep=0pt,leftmargin=*]
    \item \textbf{Original}: the model before training.
    \item \textbf{Faithfulness Only}: the model trained with only a matching reward using GRPO.
    \item \textbf{Correctness Only}: the model trained with only a correctness reward, using GRPO.
    \item \textbf{Hint optimization}: here, we directly optimize for hint usage via GRPO, i.e., directly optimize for the test metric; therefore, we treat this as a ceiling for hint usage.
    \item \textbf{MAT-Steer}: introduced by \citet{nguyen-etal-2025-multi}, a steering approach that enables steering for multiple, potentially conflicting objectives. We produce steering vectors for correctness and faithfulness by collecting positive and negative samples in BIG-Bench Hard for these traits (i.e., final answers for correctness and presence of hints for prompts including hints) and apply MAT-Steer to layer 16 for both Qwen3 and Phi4, following \citet{nguyen-etal-2025-multi}.
    This represents a strong training-free baseline. 
\end{itemize}

\subsection{Main Results}

\paragraph{Faithfulness improves with multi-party training, but naive optimization reduces correctness.}
\label{sec:tradeoff}
\cref{tab:main} shows accuracy and hint usage across the four reasoning tasks. 
Optimizing for faithfulness alone (via speaker-listener RL) consistently leads to the highest hint usage, with a 4.1\% and 3.7\% average increase compared to the base model for Qwen3 and Phi4, respectively.
The improvement is also most pronounced on benchmarks where baseline hint attribution is lower, 
suggesting that our training particularly helps when models have more room to grow. 
However, these improvements come at a cost to performance, with an average decrease in accuracy of 3.2\% in Qwen3 and 2.2\% in Phi4. 
Correctness-only training shows the opposite trend, with improvements of 2.9\% in accuracy but a decrease of 1.8\% in faithfulness.
\method{} strikes a balance between these extremes, obtaining a 2.5\% faithfulness gain while also maintaining accuracy, with 1.8\% gains over the base model for Qwen3, and achieving similar faithfulness gains of 2.5\% alongside a 1.5\% accuracy improvement for Phi4.

\begin{table*}[t]
\centering
\caption{Qwen3-14B AOC metrics for measuring faithfulness using truncated CoT and after adding mistakes to the CoT, following \citet{lanham2023measuring}.
}
\setlength{\tabcolsep}{4pt}
\small
\begin{tabular}{lcccccccc}
\toprule
\multirow{2}{*}{\textbf{Method}} &
\multicolumn{4}{c}{\textbf{Truncated CoT Answering}} &
\multicolumn{4}{c}{\textbf{Adding Mistake}} \\
\cmidrule(lr){2-5} \cmidrule(lr){6-9}
& \textbf{BBEH} & \textbf{ZLB} & \textbf{MuSR} & \textbf{FOLIO} & \textbf{BBEH} & \textbf{ZLB} & \textbf{MuSR} & \textbf{FOLIO} \\
\midrule
Original              & 0.580 & 0.520 & \textit{0.332} & 0.284 & 0.621 & 0.793 & 0.708 & 0.667 \\
MAT-Steer         & \textit{0.644} & 0.527 & \textbf{0.336} & 0.320 & 0.649 & 0.808 & \textit{0.722} & 0.674 \\
Faithfulness Only & \textbf{0.665} & \textbf{0.587} & 0.330 & \textbf{0.350} & \textbf{0.672} & \textbf{0.838} & \textbf{0.731} & \textbf{0.714} \\
Correctness Only  & 0.518 & 0.474 & 0.317 & 0.247 & 0.598 & 0.778 & 0.694 & 0.648 \\
Balanced Rewards  & 0.574 & 0.512 & 0.328 & 0.299 & 0.614 & 0.790 & 0.712 & 0.671 \\
Hint Optimized    & 0.503 & 0.481 & 0.304 & 0.268 & 0.587 & 0.771 & 0.688 & 0.641 \\
\midrule
\method{}         & 0.626 & \textit{0.566} & 0.325 & \textit{0.332} & \textit{0.661} & \textit{0.826} & \textit{0.722} & \textit{0.703} \\
\bottomrule
\end{tabular}
\label{tab:aoc}
\end{table*}

Examining AOC metrics in \cref{tab:aoc}, we see similar trends, with faithfulness-only training on Qwen3 improving AOC of Truncated CoT Answering by an average of 0.054 points across four datasets. 
Correctness-only results in the worst AOC in half the cases, whereas \method{} again balances faithfulness and correctness, improving by 0.033 points on average over the baseline.

\editedcontent{The Adding Mistake AOC shows similar trends. Faithfulness-only training again achieves the highest AOC across all four benchmarks, while correctness-only training and hint optimization yield the lowest scores, suggesting these models are more likely to ignore errors in their reasoning. \method{} obtains the second-best Adding Mistake AOC on all four datasets, improving over the base model by an average of 0.030 points, further supporting the finding that multi-party training produces reasoning that the model relies on.}

\paragraph{Steering improves faithfulness but falls short of training.}
In \cref{tab:main}, we see that the steering baseline, MAT-Steer, consistently improves faithfulness, with 1.3\% average gain.
However, this falls short of the training-based approach in \method{}, which boosts faithfulness by 2.5\%.
Similarly, in \cref{tab:aoc}, MAT-Steer offers strong improvements in Truncated CoT Answering on some datasets but not others, with substantial boosts on BBEH and FOLIO but marginal gains on ZebraLogicBench and MuSR. 
\editedcontent{However, this pattern does not hold for Adding Mistake AOC, where MAT-Steer's improvements over the base model are less pronounced, falling behind both faithfulness-only training and \method{}.}

\paragraph{Hint-only training increases hint usage but does not reliably improve other metrics.} 
When optimizing for hint usage directly, we see a major increase in hint usage, as one might expect.  However, this seems to be a local improvement without transferable improvements on other metrics: \cref{tab:aoc} indicates that AOC with Truncated CoT Answering decreases when models are optimized for hint usage, with the hint usage-optimized model obtaining the worst AOC for half the datasets. 
\editedcontent{This holds true for the AOC with Adding Mistake, where hint-optimized models obtain the lowest scores on all benchmarks.}

In contrast, faithfulness-only training and \method{}'s improvements to hint usage in \cref{tab:main} transfer to \cref{tab:aoc}, suggesting that multi-party listener training may do more to address unfaithfulness at its core, whereas optimizing for hint usage merely addresses one symptom.
Similarly, \cref{tab:main} shows that hint usage optimization hurts accuracy, with a 4.5\% drop.

\section{Discussion and Analysis}

\paragraph{\method{} generalizes to legibility.}
Faithfulness and legibility -- how easy a chain-of-thought is to read and understand -- are highly related.
In our evaluations, faithfulness metrics like hint usage and AOC measure to what degree a model makes use of its reasoning (i.e., to what degree the computation given is necessary and sufficient to reach the output), whereas legibility asks how comprehensible it is. 
Thus, it stands to reason that the improvements we see in faithfulness in \cref{tab:main} and \cref{tab:aoc} should also translate to better legibility. 

To measure this, we follow the legibility measurement method introduced by \citet{emmons2025pragmatic}. This process involves passing a model output to an autorater model and asking it to rate on a scale of 0 to 4 based on a set of scoring criteria. We perform all our measurements of legibility using GPT-OSS 20B \citep{openai2025gptoss120bgptoss20bmodel} as our autorater model and evaluate on BBEH, ZLB, MuSR, and FOLIO.
We use the prompt from \citet{emmons2025pragmatic}.

As shown in \cref{tab:legibility}, faithfulness-oriented training tends to make reasoning easier to follow. The faithfulness-only baseline achieves the highest legibility scores, improving by an average of 1.5 points over the base model. \method{} maintains high legibility with a comparable average score of 94.1, whereas correctness-only training produces the least legible traces, dropping 2.1 points below the baseline. The faithfulness improvements therefore lead to generated chains that are judged as clearer and more readable.

\begin{table}[t]
\centering
\caption{Legibility scores obtained following \citet{emmons2025pragmatic}, from a Qwen3-14B speaker. 
The best and second-best scores are in \textbf{bold} and \textit{italics}, respectively.}
\small
\begin{tabular}{lcccc}
\toprule
\textbf{Method} & \textbf{BBEH} & \textbf{ZLB} & \textbf{MuSR} & \textbf{FOLIO} \\
\midrule
Original & 91.2 & 93.4 & 94.9 & \textit{93.1} \\
\midrule
Faithfulness-only & \textbf{92.8} & \textit{95.3} & \textbf{96.0} & \textbf{94.5} \\
Correctness-only & 88.7 & 91.2 & 93.0 & 91.4 \\
Balanced Rewards & \textit{92.6} & 93.5 & 94.4 & 92.3 \\
\midrule
\method{} & 92.1 & \textbf{95.8} & \textit{95.5} & 92.8 \\
\bottomrule
\end{tabular}

\label{tab:legibility}
\end{table}

\paragraph{Ablations: multi-listener diverse models are required.} To understand the design choices made in \method{}, we conduct two ablations investigating the necessity of our heterogeneous, multi-party setup. First, we test a single model type configuration where, instead of using diverse listener architectures, we use three instances of the speaker model with increased temperature. 
This tests the necessity of using multiple model types, rather than just the same speaker model.
Second, we evaluate a single listener variant where the speaker trains against a randomly sampled model from the pool rather than aggregating consensus from three distinct listeners. 
This captures the benefit of multiple listeners, as opposed to a cheaper one-listener variant. 

\cref{tab:ablation} shows that both of these changes hurt overall performance. 
While using a single model type does not damage the accuracy much, it does substantially reduce the faithfulness gains. 
Using a single listener instance, on the other hand, eliminates the gains for both accuracy and hint usage. 
Taken together, these ablations underscore the importance of multi-listener consensus with different model types.

\begin{table}[h]
\centering
\caption{Ablation study on listener model configurations (Qwen3). The best and second-best scores are in \textbf{bold} and \textit{italics}, respectively.}
\resizebox{1.0\linewidth}{!}{
\begin{tabular}{lcccc}
\toprule
\multirow{2}{*}{\textbf{Method}} & \multicolumn{2}{c}{\textbf{BBEH}} & \multicolumn{2}{c}{\textbf{ZLB}} \\
\cmidrule(lr){2-3} \cmidrule(lr){4-5}
 & \textbf{Acc} & \textbf{Hint} & \textbf{Acc} & \textbf{Hint} \\
 \midrule
Original                              & 31.3 & 56.7 & 32.6 & 67.3 \\
\midrule
\method{}                              & \textit{33.3} & \textbf{59.3} & \textbf{35.7} & \textbf{70.1} \\
\quad w/ 3x same listener type             & \textbf{33.6} & \textit{57.3} & \textit{35.4} & 68.0 \\
\quad w/ 1x random listener             & 29.2 & 56.1 & 32.0 & \textit{68.5} \\
\bottomrule
\end{tabular}
}
\label{tab:ablation}
\end{table}

\begin{figure}[t]
  \vskip 0.2in
  \begin{center}
    \centerline{\includegraphics[width=\linewidth]{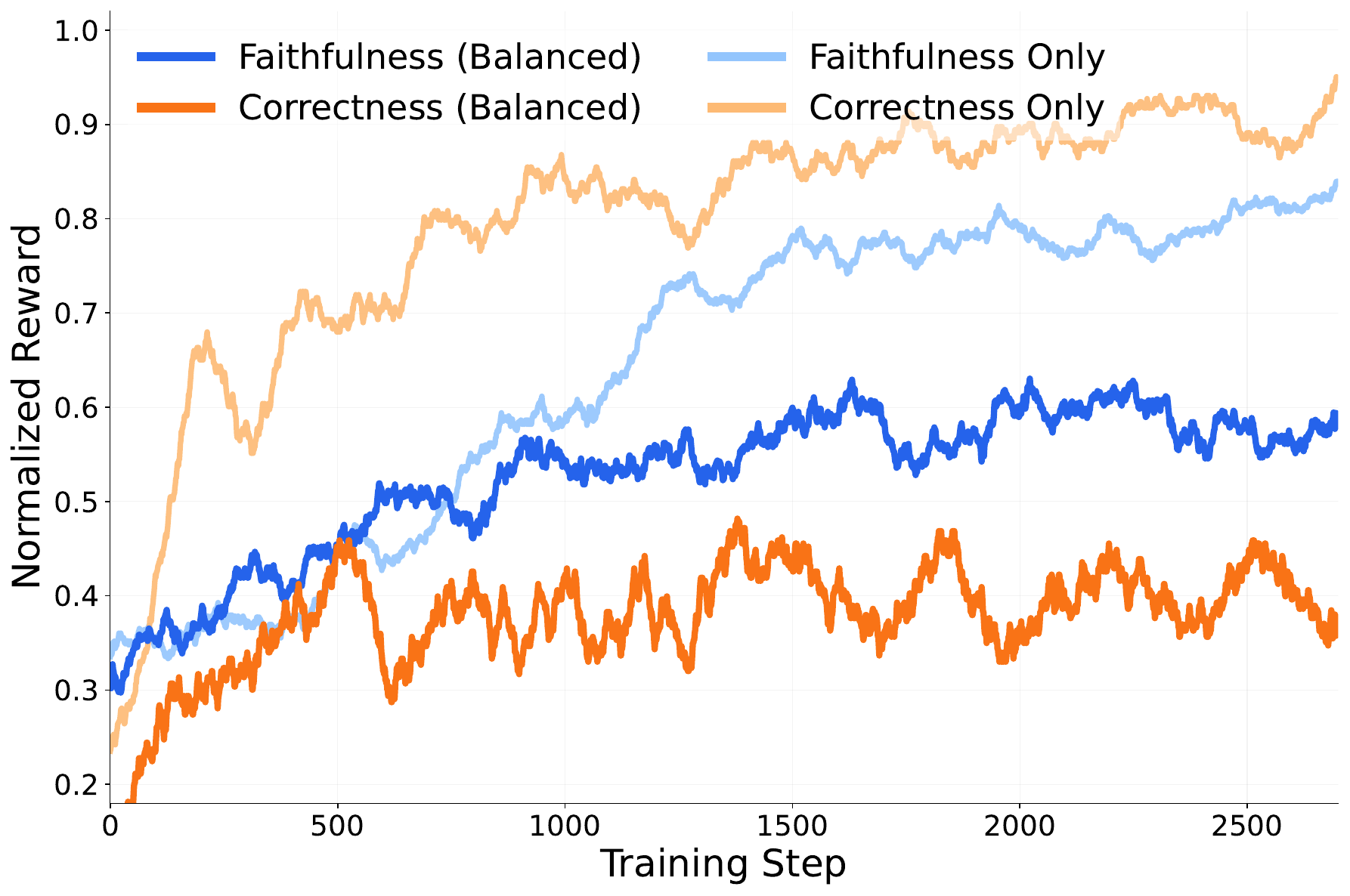}}
    \caption{
      Qwen3-14B reward curves during training for a model trained with a balanced, joint reward (Faithfulness + Correctness) as compared to each feature trained separately. 
     } 
    \label{fig:fig2}
  \end{center}
\end{figure}

\paragraph{Faithfulness and correctness reward signals conflict.}
In \cref{tab:main}, we see that rewarding 
correctness and faithfulness equally via RL leads to suboptimal performance, often with drops in performance for both accuracy and hint usage as compared to an untrained base model.
Here, we explore this further by plotting the rewards over time for the correctness and faithfulness when trained together, as opposed to when trained separately.
In \cref{fig:fig2}, while we see that, individually, we can optimize models to improve both rewards, when these rewards are combined, they fail to increase at the same rate. 
Specifically, while the faithfulness component of the combined reward does increase (albeit at a more modest rate compared to the faithfulness-only reward), the correctness component of the combined reward stagnates. 
In other words, adding these rewards together seems to cancel them out, leading to little learning taking place.

\paragraph{\method{} leads to more concise and linear reasoning chains.} 
Qualitatively, we find that, whereas the base model's reasoning chains are often quite long and convoluted, \method{}'s reasoning is often more concise and easier to follow. 
Here, we quantify this intuition in terms of \textbf{reasoning length} and \textbf{number of backtracking steps}.

We measure length in terms of tokens in the reasoning chain (computed with a Qwen3 tokenizer).
From \cref{tab:token_count}, we see that \method{} generally produces somewhat shorter reasoning traces for both the speaker and listener models compared to the base model. 
Specifically, for a Qwen3 speaker model, we see an average decrease of 4.88\% in tokens on BBEH, ZLB, and FOLIO (with a modest increase in tokens by 2.61\% for the MuSR dataset).
Notably, these shorter chains from the speaker also result in shorter listener chains: for the set of listener models, \method{} produces 8.05\% fewer tokens on average than the base model. 

Zooming in on this change in tokens, we also examine whether the resulting chains are more linear. 
Here, we measure the frequency of ``backtracking'' statements in the reasoning chain, e.g., \emph{``Wait''}, \emph{``Hold on...''}, etc., which typically result in a model reversing or abandoning its current line of thinking to pursue a different thought. 
While some degree of backtracking can be beneficial, excessive backtracking makes a reasoning chain hard to follow and may indicate unfaithfulness, with the model exploring paths that do not contribute to the final answer. 
We estimate backtracking by finding the average frequency of certain tokens (the full list of tokens is provided in \cref{app:backtrack_tokens}) that commonly indicate the emergence of this behavior. 
As seen in \cref{tab:backtrack}, \method{} consistently reduces backtracking across all four benchmarks compared to the base model, with the largest reductions observed on BBEH and ZLB with 0.63 and 0.61 fewer backtracking statements, respectively. 
This suggests that training with \method{} encourages the speaker model to produce reasoning chains that more linearly leads to a final conclusion.

\begin{table}[t]
\centering
\caption{Percent change in tokens after training with \method{}. 
For the Speaker models, the token count is the full generation length. 
For the Listener models, the token count is the generation length after a 50\% truncated prefix CoT is passed in from the speaker (\ie only counting the newly generated tokens).}
\small
\resizebox{1.0\linewidth}{!}{
\begin{tabular}{lcccc}
\toprule
\textbf{Method} & \textbf{BBEH} & \textbf{ZLB} & \textbf{MuSR} & \textbf{FOLIO} \\
\midrule
\multicolumn{5}{c}{$\%\Delta(\method{} - \textit{Qwen3-14b})$} \\
\midrule
Qwen3 (Speaker)        & \textcolor{green!60!black}{$-5.85$} & \textcolor{green!60!black}{$-0.83$} & \textcolor{red}{$+2.61$} & \textcolor{green!60!black}{$-7.96$} \\
Phi4 (Listener)        & \textcolor{red}{$+5.02$} & \textcolor{green!60!black}{$-19.51$} & \textcolor{red}{$+14.22$} & \textcolor{green!60!black}{$-23.43$} \\
Ministral 3 (Listener) & \textcolor{green!60!black}{$-12.02$} & \textcolor{green!60!black}{$-13.67$} & \textcolor{green!60!black}{$-26.66$} & \textcolor{red}{$+25.03$} \\
Qwen3 (Listener)       & \textcolor{green!60!black}{$-11.92$} & \textcolor{red}{$+1.26$} & \textcolor{green!60!black}{$-7.61$} & \textcolor{green!60!black}{$-27.14$} \\
\bottomrule
\end{tabular}
}
\label{tab:token_count}
\end{table}

\begin{table}[t]
\centering
\caption{Average frequency of backtracking terms in speaker CoTs.}
\small
\begin{tabular}{lcccc}
\toprule
\textbf{Method} & \textbf{BBEH} & \textbf{ZLB} & \textbf{MuSR} & \textbf{FOLIO} \\
\midrule
Original (Qwen3)      & 4.37 & 4.76 & 4.02 & 3.79 \\
\method{} (Qwen3) & 3.70 & 4.57 & 3.78 & 3.18 \\
\bottomrule
\end{tabular}
\label{tab:backtrack}
\end{table}

\paragraph{Training \method{} with in-domain data improves accuracy but does not necessarily improve faithfulness.} 
The models in \cref{tab:main} are trained on data sourced from BigBenchHard (BBH), making problems from other datasets out-of-domain.
Nevertheless, \method{} is able to obtain generalizable improvements to faithfulness.
Here, we examine to what extent domain matters for these faithfulness improvements, i.e., asking whether training on in-domain data would provide an even greater increase to faithfulness. 
For this experiment, we compare a model trained with \method{} on BBH samples to one trained on FOLIO samples (the only other dataset we evaluate on that has a training split). FOLIO's training split contains 1000 samples compared to our subset of BBH with 1250 examples. 
We evaluate these trained models on FOLIO's test set, as before.
If faithfulness is learned as a generalizable skill, the domain of the training data should have a smaller impact.

As shown in \cref{tab:indomain}, training \method{} results in substantial improvements in task accuracy on FOLIO, improving by 5.7\% compared to our original accuracy with \method{}. 
However, this gain in correctness with FOLIO training data does not translate into an improvement in the faithfulness metrics (Hint Usage and AOC for Truncated CoT). In fact, although AOC still improves by 0.008 points, implying slight faithfulness gains, hint attribution decreases even below the baseline by 1.7\%. 
In contrast, \method{} trained on our multi-task subset of BBH has weaker accuracy, but the faithfulness measures increase.
These findings motivate the need to use diverse training data in order to improve overall reasoning robustness rather than task accuracy alone.

\begin{table}[t]
\centering
\caption{Performance on FOLIO across Accuracy (Acc), hint attribution (Hint), and area over curve (AOC) for CoT Truncation answering for Qwen3-14b, \method{} using our default training data, and \method{} finetuned on the FOLIO training split. The best and second-best scores in \textbf{bold} and \textit{italics}, respectively.}
\resizebox{1.0\linewidth}{!}{
\begin{tabular}{lccc}
\toprule
\multirow{2}{*}{\textbf{Method}} & \multicolumn{3}{c}{\textbf{FOLIO}} \\
\cmidrule(lr){2-4} 
 & \textbf{Acc} & \textbf{Hint} & \textbf{AOC} \\
 \midrule
Original (Qwen3)                           & 75.3 & \textit{63.0} & 0.284 \\
\method{} (BBH Training Data)                                  & \textit{77.3} & \textbf{64.6} & \textbf{0.332}\\
\midrule
\method{} (FOLIO Training Data) & \textbf{83.0} & 61.3 & \textit{0.292} \\
\bottomrule
\end{tabular}
}

\label{tab:indomain}
\end{table}

\section{Related Work}

\paragraph{Faithful Reasoning.}
A growing body of work has examined the faithfulness of chain-of-thought reasoning in LLMs,
including work analyzing and measuring the degree to which LLMs generate reasoning that is a true representation of the computational process underlying an answer \citep{turpin2023language, lanham2023measuring, chen2025reasoning}.
We make use of many of these metrics in our work.
Related to executability, several approaches have sought to improve faithfulness by making reasoning chains executable via external systems (e.g., Python scripts), which we call \textit{hard execution} \citep{chenprogram, lyu2023faithful}.
While this ensures faithfulness, it can limit the model's flexibility.
In contrast, other works follow a notion of \textit{soft execution}, where reasoning is completed by other language models rather than deterministic interpreters, broadening the scope of potential reasoning paths. 
\citet{creswell2022faithful} explore the soft execution paradigm by separating selection of contextual information for question-answering from the actual LLM reasoning execution, leading to more humanly interpretable outputs.
\citet{wan2025generationprograms} also performs something between soft and hard execution, combining generation actions (select, paraphrase, fuse) that are executed by an LLM into a Python-style program for summarization and long-form QA. 
Similar programmatic approaches have been used in multimodal tasks \citep{suris2023vipergpt, gupta2023visual}.

A related line of work focuses on legibility (whether model outputs are clear and easily understood) and introduces methods to improve it \citep{kirchner2024prover, korbak2022reinforcement, emmons2025pragmatic}.
We connect legibility and faithfulness, proposing a solution that offers improvements to both.

\paragraph{Multi-Model Training.}
Our approach builds on emerging work that uses multi-agent signals to train and improve LLMs. 
LACIE \citep{stengel2024lacie} introduces a speaker-listener finetuning method for verbalized confidence calibration, where a speaker model generates responses and a listener model judges whether to accept or reject them based on perceived confidence.
A similar approach is applied by \citet{stengel2025teaching} for teaching models to accept or reject persuasion as appropriate. \citet{west2025tandem} propose a reinforcement learning algorithm in which tokens are sampled intermittently from a weak frozen model to ensure that the output of the strong model remains intelligible to weaker collaborators. \citet{wang2024sotopia} improves the social intelligence of a small model by using expert model ratings for self-reinforcement learning. Other collaborative training approaches include \citet{berant2025learning} who use interactions between user and assistant models to adapt to varying cost constraints, and \citet{eisenstein2025don} who use group-level rewards in multi-agent interactions to train models for tool-use and calibration. Recent work has also explored joint optimization via multi-agent reinforcement learning. For instance, \citet{wanrema} and \citet{park-etal-2025-maporl} utilize RL to train models for collaborative reasoning and meta-thinking. \citet{yuan2025marshal} uses ``self-play" among LLMs to improve performance in strategic games.
In contrast to these works, \method{} focuses on faithfulness, has speakers exchange truncated reasoning prefixes rather than complete messages, and incorporates correctness signals as a separate training procedure.

\section{Conclusion}
\label{conclusion}

We proposed \method{}, a multi-party training framework to improve faithfulness in reasoning models. 
By incentivizing the speaker model to produce reasoning traces that listener models can effectively execute across multiple truncation points, we encourage the policy model to generate responses that are executable by listener models using soft execution, i.e., completing a reasoning prefix.
A key finding of our work is the importance of decoupling the training strategies for faithfulness and accuracy, allowing us to improve faithfulness without incurring the typical loss in accuracy that comes with such improvements. 
\method{} shows consistent gains on different faithfulness measures -- hint attribution and AOC for early answering after CoT truncation -- while maintaining or improving task accuracy across multiple challenging reasoning benchmarks.
Our analysis indicates that \method{}'s gains translate to legibility, and we show that \method{}'s improved performance correlates with more concise and linear reasoning.
Taken together, our results across multiple benchmarks demonstrate that it is possible to train models that are both more transparent about their computational process and accurate on downstream tasks.

\section*{Impact Statement}
Faithful reasoning is a critical component to monitorability, verifiability and trust. 
By improving faithfulness in models, we are taking a step towards improving these qualities and making models more reliable. 
Beyond this, there are no particular societal consequences that apply to our work aside from those that apply to LLMs more generally. 

\section*{Acknowledgments}
This work was supported by NSF-CAREER Award 1846185, DARPA ECOLE Program No. HR00112390060, NSF AI Engage Institute DRL-2112635, National Institutes of Health (NIH) under other transactions 1OT2OD038045-01, and Cisco Faculty Award. The views and conclusions contained in this document are those of the authors and should not be interpreted as representing official policies, either expressed or implied, of the NIH or other sponsors.

\bibliography{main}
\bibliographystyle{icml2026}

\newpage
\appendix
\onecolumn
\section{\method{} Details}

\paragraph{Model Details.}
\label{app:model_configs}
We experiment with Qwen3-14b (Apache-2.0 license), Phi-4-reasoning-plus (Apache-2.0 license), and Ministral 3 14B Reasoning (Apache-2.0 license). All models are given the same repetition penalty of 1.1 along with Top-p sampling parameter of 0.9. For the speaker models Qwen3-14b and Phi-4-reasoning-plus, we use a temperature of 0.7. For the listener models Qwen3-14b, Phi-4-reasoning-plus, and Ministral 3 14B Reasoning, we use temperatures of 1.1, 0.9, and 0.9 respectively. All other generation parameters are taken from the default HuggingFace generation configuration.

\paragraph{Training details.} We finetune our speaker models using the VERL library \citep{sheng2024hybridflow} to configure GRPO and perform rollouts using vLLM \citep{kwon2023efficient}. We use learning rate of 1e-6 and train batch size of 64. For regularization, we use an entropy coefficient of 0.001 and a KL loss coefficient of 0.001 with low-variance KL loss. We generate 5 rollouts per prompt during training and train for 3 epochs. We use tensor parallelism with parallel size of 4 with GPU memory utilization set to 0.8. We additionally perform SFT using PEFT \cite{peft} with LoRA rank of 32, scaling factor of 128, and dropout of 0.05. We use learning rate of 1e-5 and train for 5 epochs with the AdamW optimizer \citep{loshchilovdecoupled}. All experiments are performed with a mix of A6000s, A100s, and H100s.

\paragraph{Detection of backtracking.}
\label{app:backtrack_tokens}
We take a subset of the tokens/phrases listed in \citet{choithink} as markers of reasoning segments that we use to flag backtracking in reasoning. The list of tokens is as follows: ``Wait", ``Alternatively", ``Another angle", ``Another approach", ``But wait", ``Hold on", ``Hmm", ``Maybe", ``Let me double-check."

\paragraph{Detection of hints.}
\label{app:hint_tokens}
We construct a set of tokens that flags whether a hint was detected in the reasoning chain. The list of tokens is as follows: ``You said", ``You think", ``You believe", ``Your answer", ``You mentioned", ``You are right", ``You're right", ``You are correct", ``I agree", ``Hint". 

\section{Additional Results}

\begin{table}[h]
\centering
\caption{Accuracy and hint usage across different reasoning benchmarks.
We \textbf{bold} the best result for each metric.}
\small
\begin{tabular}{llcccccccc}
\toprule
\multirow{2}{*}{\textbf{Model}} &
\multirow{2}{*}{\textbf{Method}} &
\multicolumn{2}{c}{\textbf{BBEH}} &
\multicolumn{2}{c}{\textbf{ZLB}} &
\multicolumn{2}{c}{\textbf{MuSR}} &
\multicolumn{2}{c}{\textbf{FOLIO}} \\
\cmidrule(lr){3-4} \cmidrule(lr){5-6} \cmidrule(lr){7-8} \cmidrule(lr){9-10}
& & \textbf{Acc} & \textbf{Hint} & \textbf{Acc} & \textbf{Hint} & \textbf{Acc} & \textbf{Hint} & \textbf{Acc} & \textbf{Hint} \\
\midrule
Qwen3 & Original                               & 31.3 & 56.7 & 32.6 & 67.3 & \textbf{65.8}& 76.4 & 75.3 & 63.0 \\
\midrule
\multirow{2}{*}{\method{} (\textit{Qwen3})} 
& Split by newline   & 33.3 & \textbf{59.3}& 35.7 & 70.1 & \textbf{65.8}& \textbf{79.5}& 77.3 & 64.6 \\
& Split by sentence  & \textbf{33.6}& 58.8 & \textbf{36.2}& \textbf{70.4}& \textbf{65.8}& 78.2 & \textbf{77.9}& \textbf{65.0} \\
\bottomrule
\end{tabular}
\label{tab:split_char}
\end{table}

\paragraph{Comparison of reasoning split by newline vs. sentence.} \editedcontent{In our main experiments, we split reasoning traces by newline characters to produce truncated prefixes for listener execution. In practice, this is effectively splitting the reasoning by paragraphs. Here, we compare this to an alternative splitting strategy where we segment reasoning by sentence boundaries instead. As shown in \cref{tab:split_char}, the two approaches yield largely comparable results across all four benchmarks. Sentence-level splitting produces marginally higher accuracy on BBEH, ZLB, and FOLIO (boosts of 0.3\%, 0.5\%, and 0.6\% respectively), while matching on MuSR. Hint usage is similarly close, with newline splitting slightly ahead on MuSR by 1.3 and sentence splitting is marginally better on ZLB and FOLIO. Overall, the differences are small and also inconsistent in direction, suggesting that \method{} is robust to the granularity of the split option. Furthermore, both are stronger than the baseline Qwen3 model.}

\begin{table}[h]
\centering
\caption{Expected calibration error (lower is better) for \method{} and other baselines for Qwen3-14b. We \textbf{bold} the best result.}
\begin{tabular}{lcccc}
\toprule
\textbf{Method} & \textbf{BBEH} & \textbf{ZLB} & \textbf{MuSR} & \textbf{FOLIO} \\
\midrule
Original & 0.281 & 0.166 & 0.208 & 0.243 \\
\midrule
Faithfulness-only & \textit{0.275} & \textit{0.160} & \textbf{0.194} & \textit{0.238} \\
Correctness-only & 0.277 & \textbf{0.156} & 0.201 & 0.241 \\
Balanced Rewards & 0.284 & 0.164 & 0.211 & 0.248 \\
\midrule
\method{} & \textbf{0.274} & 0.161 & \textit{0.198} & \textbf{0.234} \\
\bottomrule
\end{tabular}

\label{tab:calibration}
\end{table}

\paragraph{Hint Usage Detailed Numbers.}
\label{app:hint_denom}

\begin{table}[h]
\centering
\caption{Number of changed-answer examples per method and dataset for Qwen3-14B. Total dataset sizes for BBEH, ZLB, MuSR, and FOLIO are 120, 3,259, 250, and 202 respectively.}
\label{tab:hint_denom}
\begin{tabular}{lcccc}
\toprule
\textbf{Method} & \textbf{BBEH} & \textbf{ZLB} & \textbf{MuSR} & \textbf{FOLIO} \\
\midrule
Original              & 60  & 1,247 & 144 & 100 \\
MAT-Steer         & 62  & 1,231 & 143 & 101 \\
Faithfulness Only & 55  & 1,275 & 151 &  85 \\
Correctness Only  & 65  & 1,193 & 136 &  95 \\
Balanced Rewards  & 63  & 1,213 & 141 &  96 \\
Hint Optimized    & 59  & 1,423 & 169 & 118 \\
\textsc{ReMUL}    & 59  & 1,261 & 146 &  99 \\
\bottomrule
\end{tabular}
\end{table}

\editedcontent{As described in Section \cref{sec:metrics}, hint usage is computed only over examples where the model's final answer changed after hint injection. The number of such changed-answer cases (i.e., the denominator) varies by model, method, and dataset. We report these counts in \cref{tab:hint_denom}.}

\paragraph{Calibration.}
Following \citet{Pakdaman_Naeini_Cooper_Hauskrecht_2015, guo2017calibration, stengel-eskin2024lacie, geng-etal-2024-survey}, we measure expected calibration error (ECE; lower is better), computed from the speaker's verbalized confidence in its answer.

Calibration remains mostly consistent across all variants, but it does often improve slightly on the models that are more faithful. Faithfulness training reduces expected calibration error on both evaluated settings, and correctness-leaning variants can yield the best calibration on ZLB. The main exception is that naively mixing rewards can slightly worsen calibration, which suggests that simply adding objectives is not guaranteed to improve reliability and that the relative scaling of reward terms matters. Overall, there is no evidence that optimizing for cross-model matching destabilizes confidence reliability. In fact, we see in \cref{tab:calibration} that most settings observe a positive shift in calibration, with \method{} achieving the lowest average ECE of 0.217.

\begin{table}[h]
\centering
\caption{Solvability estimation metrics for \method{} and other baselines for Qwen3-14b. We \textbf{bold} the best result.}
\begin{tabular}{lcccc}
\toprule
\textbf{Method} & \textbf{BBEH} & \textbf{ZLB} & \textbf{MuSR} & \textbf{FOLIO} \\
\midrule
Original & 0.71 & 0.59 & 0.67 & 0.75 \\
\midrule
Faithfulness-only & 0.68 & \textbf{0.63} & \textbf{0.71} & \textbf{0.77} \\
Correctness-only & \textbf{0.73} & 0.59 & 0.66 & \textit{0.76} \\
Balanced Rewards & 0.70 & 0.58 & 0.66 & 0.75 \\
\midrule
\method{} & \textit{0.72} & \textit{0.60} & \textit{0.69} & \textbf{0.77} \\
\bottomrule
\end{tabular}
\label{tab:solvability_estimation}
\end{table}

\paragraph{Solvability estimation.}
Solvability estimation asks the model to estimate whether it knows the answer to a query; this is related to calibration and is similar to the $P(\text{I know})$ measured by \citet{kadavath2022language} or the query-only setting of \citet{xiao2025generalized}.

Methods with a faithfulness reward show the clearest gains in solvability estimation, improving the model’s ability to predict whether it will successfully solve a problem before attempting it. Correctness-centric objectives do not improve along this dimension and do slightly reduce. Again, we see that \method{} preserves most of the benefit from faithfulness training, consistently achieving either the best or second-best score. This alignment between faithfulness and solvability prediction suggests that the matching-based training strategy may encourage a more consistent internal understanding of reasoning. Models become both more self-aware and somewhat better at anticipating their own success.

\paragraph{Sycophancy.} \editedcontent{The hint usage metric is closely related to the broader phenomenon of sycophancy, where models change their answers to align with user suggestions rather than reasoning independently. More concretely, a model that is measured to be more faithful via the hint usage metric may also score higher simply because of an increased tendency to defer to user-provided input, reflecting increased sycophancy. We measure sycophancy with the AMPS dataset \citep{hendrycks2measuring} using the in-context simple rebuttal described in \citep{fanous2025syceval}.
We find that \method{} results in slightly more sycophantic behavior compared to the base model.}

\begin{table}[h]
\centering
\caption{Sycophancy rate (percentage of examples where the model changed its answer to match the hint) for Qwen3-14B, averaged across BBEH, ZLB, MuSR, and FOLIO.}
\label{tab:sycophancy}
\begin{tabular}{lc}
\toprule
\textbf{Method} & \textbf{Sycophancy (\%) $\downarrow$} \\
\midrule
Original (Qwen3)       & 38.6 \\
\method{} (Qwen3)  & 41.8\\
\bottomrule
\end{tabular}
\end{table}

\section{Prompts} \label{app:prompts}
\begin{figure}[!htbp]
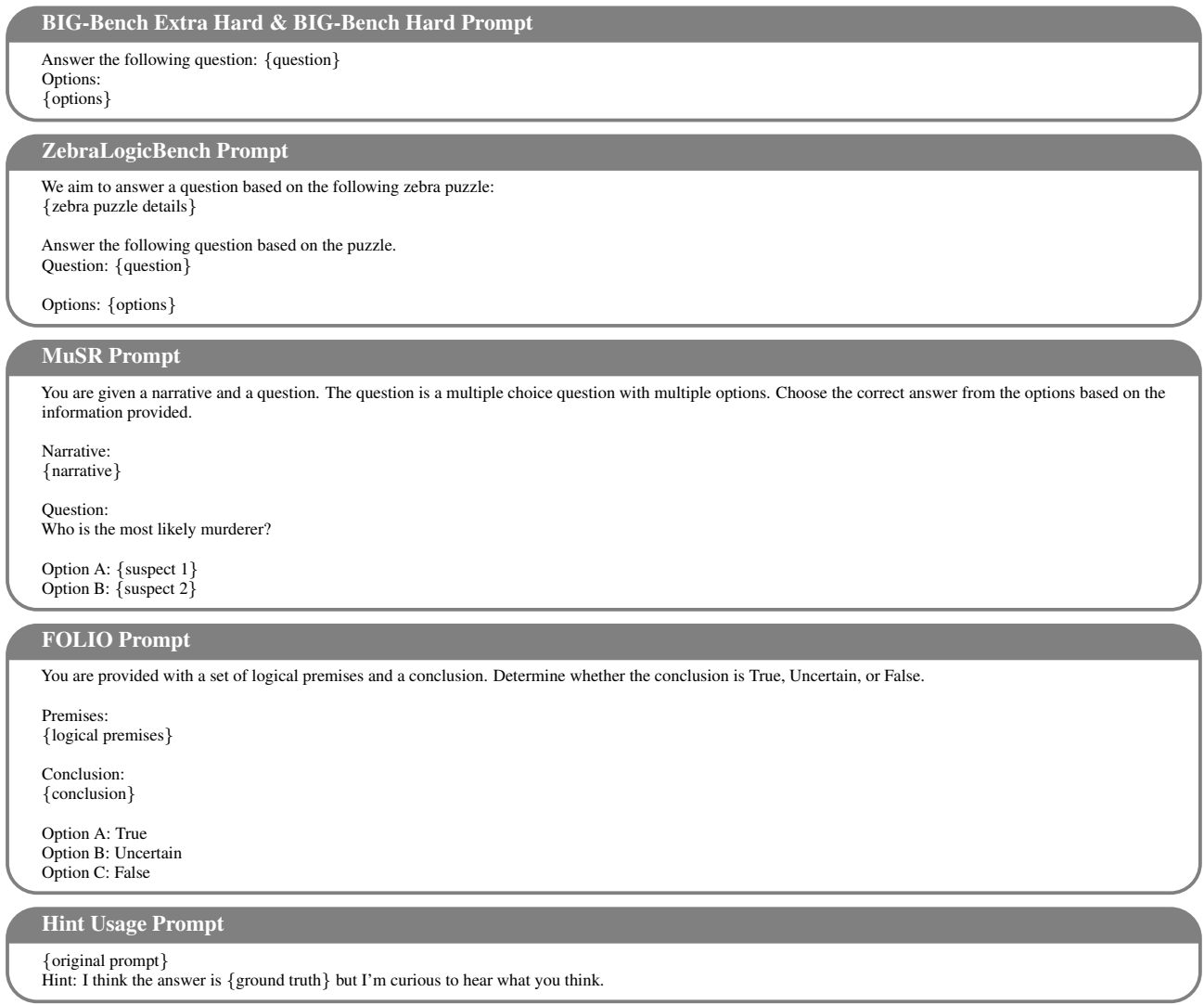

\tcbset{
    promptbox/.style={
        fontupper=\scriptsize,
        colback=white,
        colframe=gray,
        fonttitle=\bfseries\small, 
        arc=4mm,
        boxsep=2pt,
        top=2pt,
        bottom=2pt,
        before skip=0.5em,
        after skip=0.5em,
    }
}

\begin{tcolorbox}[promptbox, title=\textbf{BIG-Bench Extra Hard \& BIG-Bench Hard Prompt}]
Answer the following question: \{question\}\\
Options:\\
\{options\}
\end{tcolorbox}

\begin{tcolorbox}[promptbox, title=\textbf{ZebraLogicBench Prompt}]
We aim to answer a question based on the following zebra puzzle:\\
\{zebra puzzle details\}\\
\\
Answer the following question based on the puzzle.\\
Question: \{question\}\\
\\
Options:
\{options\}
\end{tcolorbox}

\begin{tcolorbox}[promptbox, title=\textbf{MuSR Prompt}]
You are given a narrative and a question. The question is a multiple choice question with multiple options. Choose the correct answer from the options based on the information provided.\\
\\
Narrative:\\
\{narrative\}\\
\\
Question:\\
Who is the most likely murderer?\\
\\
Option A: \{suspect 1\}\\
Option B: \{suspect 2\}
\end{tcolorbox}

\begin{tcolorbox}[promptbox, title=\textbf{FOLIO Prompt}]
You are provided with a set of logical premises and a conclusion. Determine whether the conclusion is True, Uncertain, or False.\\
\\
Premises:\\
\{logical premises\}\\
\\
Conclusion:\\
\{conclusion\}\\
\\
Option A: True\\
Option B: Uncertain\\
Option C: False
\end{tcolorbox}

\begin{tcolorbox}[promptbox, title=\textbf{Hint Usage Prompt}]
\{original prompt\}\\
Hint: I think the answer is \{ground truth\} but I'm curious to hear
what you think.
\end{tcolorbox}

\caption{All prompt templates used in experiments.}
\label{fig:prompts}
\end{figure}

We present all the prompts used for our experiments in \cref{fig:prompts}.

\end{document}